\documentclass[sigconf]{acmart}

\usepackage{booktabs} 
\usepackage{url}
\usepackage{hyperref}
\usepackage{balance}

\begin{document}

\title{Deep Cross Modal Learning for Caricature Verification and Identification(CaVINet)}

\author{Jatin Garg}\authornote{Authors with equal contribution}
\affiliation{
    \institution{Indian Institute of Technology Ropar}
}
 \email{2014csb1017@iitrpr.ac.in}
 
\author{Skand Vishwanath Peri}\authornotemark[1]
\affiliation{
    \institution{Indian Institute of Technology Ropar}
}
  \email{pvskand@gmail.com}
  
\author{Himanshu Tolani}\authornotemark[1]
\affiliation{
    \institution{Indian Institute of Technology Ropar}
}
  \email{2014csb1015@iitrpr.ac.in}
  
\author{Narayanan C Krishnan}
\affiliation{
    \institution{Indian Institute of Technology Ropar}
}
  \email{ckn@iitrpr.ac.in}

\subtitle{\href{https://lsaiml.github.io/CaVINet/}{\texttt{\textcolor{red}{https://lsaiml.github.io/CaVINet/}}}}

\renewcommand{\shortauthors}{Jatin Garg, Skand V Peri, Himanshu Tolani, and Narayanan C Krishnan}

\begin{abstract}
Learning from different modalities is a challenging task. In this paper, we look at the challenging problem of cross modal face verification and recognition between caricature and visual image modalities. Caricature have exaggerations of facial features of a person. Due to the significant variations in the caricatures, building vision models for recognizing and verifying data from this modality is an extremely challenging task. Visual images with significantly lesser amount of distortions can act as a bridge for the analysis of caricature modality. We introduce a publicly available large \textbf{Ca}ricature-\textbf{VI}sual  dataset [CaVI] with images from both the modalities that captures the rich variations in the caricature of an identity. This paper presents the first cross modal architecture that handles extreme distortions of caricatures using a deep learning network that learns similar representations across the modalities. We use two convolutional networks along with transformations that are subjected to orthogonality constraints to capture the shared and modality specific representations. In contrast to prior research, our approach neither depends on manually extracted facial landmarks for learning the representations, nor on the identities of the person for performing verification. The learned shared representation achieves 91\% accuracy for verifying unseen images and 75\% accuracy on unseen identities. Further, recognizing the identity in the image by knowledge transfer using a combination of shared and modality specific representations, resulted in an unprecedented performance of 85\% rank-1 accuracy for caricatures and 95\% rank-1 accuracy for visual images.
\end{abstract}

%
%


\copyrightyear{2018}
\acmYear{2018}
\setcopyright{acmcopyright}
\acmConference[MM '18]{2018 ACM Multimedia Conference}{October 22-26, 2018}{Seoul, Republic of Korea}

\acmPrice{15.00}
\acmDOI{10.1145/3240508.3240658}
\acmISBN{978-1-4503-5665-7/18/10}
\settopmatter{printacmref=true} 
\maketitle
\fancyhead{}
\section{Introduction}

 Deep learning has been quite effective in narrowing the representational gap for cross modal learning applications such as cross modal face recognition/verification. Prior work in this area focused on multi-modal facial images such as near-infrared, forensic sketches, depth imagery, etc. Approaches for these modalities have been quite successful due to the inherent similarity in the structure of a face captured using different modalities. However, cross-modal face analytics where one of the modality is a caricature is a challenging task due to the extreme levels of distortions. 



\begin{figure}[h]
\centering
\includegraphics[width=0.4\textwidth]{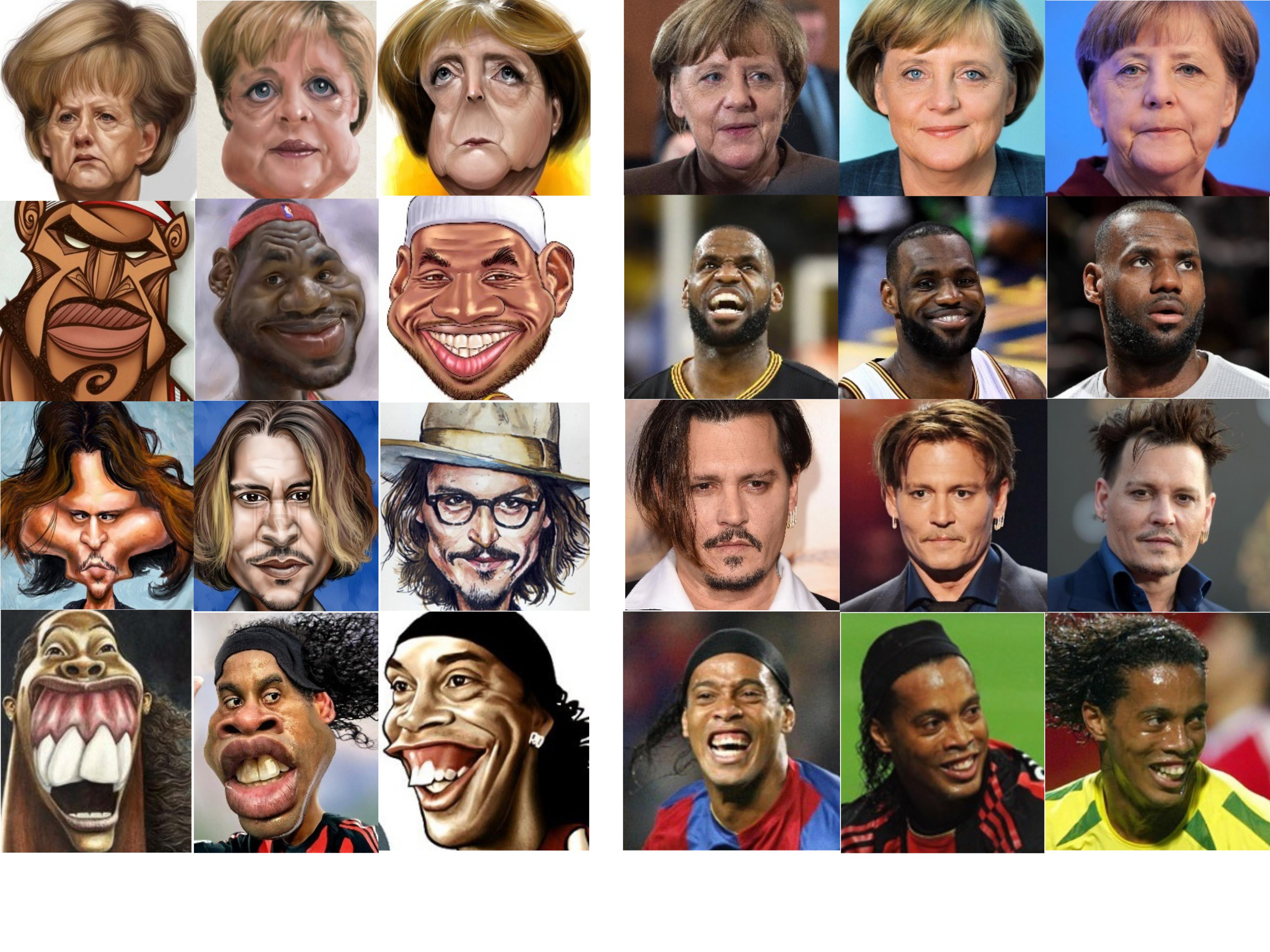}
\caption{Caricature and visual image examples from the CaVI dataset.}.
\label{fig:dataset}
\end{figure}

Caricatures are drawings with extreme distortion of a person's facial features. A caricature is not a sketch because a sketch preserves facial structure and features to a large extent. A caricature on the other hand may have variations with respect to the expressions, point of view, appearance, and also the underlying artistic style. Figure \ref{fig:dataset} illustrates caricatures of few identities, where the distortions are apparent. In spite of the myriad distortions in a caricature, humans are adept in recognizing the identity of the caricature as well as verifying whether the caricature and a visual image of a face correspond to same or different identities. 

While deep learning models have been successful at face verification and recognition in the wild \cite{Facenet, Parkhi15}, the same cannot be said with respect to caricatures. The difficulty stems from the exaggerations and distortions present in the caricatures that vary significantly from one image to another. Caricatures being distorted views of real faces still posses some distinctive characteristics that assist humans to verify and recognize the identities in the images. These distinctive characteristics present in both the modalities if captured accurately can aid in caricature verification and identification tasks. 

In this paper, we address the two problems of caricature verification and identification. Caricature-visual verification refers to the task of verifying whether the two input images of different modalities(one each of caricature and visual modality) correspond to the same identity. Caricature recognition aims at identifying the person in the caricature image. We present the first approach that gives promising results for these two tasks. 

We propose and engineer CaVINet, a novel coupled-deep neural network model that captures the shared representations between the modalities for performing the verification task, and utilizes the shared and modality-specific representations for performing the identification task. The coupled network consists of individual branches for the two modalities that are connected to perform the two tasks simultaneously. The connection between the two branches is constrained to explicitly capture shared and modality specific representations. In contrast to prior work, CaVINet does not share all the parameters of the modality specific branches to facilitate the characterization of meaningful representations. Further, it explicitly models the verification task, rather than performing verification through identification. This allows CaVINet to perform the verification task on identities that were not part of the training set. We achieve 91\% accuracy for the verification task and 85\% accuracy for caricature identification task. By coupling the visual and caricature networks we observe successful transfer of knowledge from visual to caricature modality, resulting in a boost in the performance of caricature identification. We also introduce a new publicly available multimedia dataset, caricature and visual images (CaVI) for caricature analysis. Over all, the proposed work makes the following contributions:
\begin{itemize}
\item We present the first facial key point less approach for caricature identification and verification. Prior approaches for cross-modal verification of caricatures \cite{rw1, rw5} require manually extracted facial key points for aligning the cross-modal images, a difficult task due to distortions in the facial features. 
In contrast, the proposed solution offers a facial key point less method for caricature verification and identification. 
The proposed CaVINet model successfully handles extreme distortions present in the caricatures by transferring knowledge from the visual modality, as observed from the high accuracy for cross modal verification and identification.
\item We introduce a new publicly available dataset (CaVI) that contains caricatures and visual images of 205 identities. The dataset is diverse and has significant variations in terms of view points, facial expressions, artistic imagination, and exaggerated facial feature in the caricatures of an identity. Further, the dataset also provides the manually annotated bounding box of the face in a caricature. 
\item CaVINet is a generic model for cross modal caricature verification without requiring the identities of the test samples being part of the train set. This is a significant change over the current approaches \cite{rw1, rw5}, which perform verification by identification, thereby requiring the test identities to be part of training set as well. 
\end{itemize}

\section{Related Work}

While traditional face recognition has been explored to a large extent, there has been limited work on caricature identification and verification. All the caricature identification approaches to date either require manual extraction of features from the images \cite{rw1, rw2, rw3}, or manual definition of facial key points from the images. The oldest work on caricature identification is by Klare et al., \cite{rw1} who explore various machine learning models on a small dataset containing 196 pairs of face images and caricatures. Hospedales et al., \cite{rw2} propose a method that trains SVM classifiers on manually extracted features combined with low-level features extracted using histogram of gradients, local binary patterns followed by principal component analysis for recognition using canonical correlation analysis (CCA). Abaci and Akgul\cite{rw3} train SVM on features extracted from an extremely small dataset consisting of 200 pairs of caricatures and visual images. This is followed by a genetic algorithm and logistic regression to find optimum weights which reduces the distance between caricature and photograph features. 



There has been some recent work on caricature verification and identification using features extracted from manually annotated facial key points  \cite{rw5} \cite{newWebCari_Paper}. Huo et al., \cite{rw5} propose to align the caricature and visual images using facial key points. Off the shelf VGGBox model was used to extract features from the aligned images. Kernel coupled spectral regression model trained on the extracted features resulted in the best performance of 65\% accuracy for verification and 55\% accuracy for identification tasks. These results were obtained on the WebCaricature dataset. The primary drawback of this approach is the manual extraction of facial key points from caricatures for aligning the images of different modalities. Huo et al., \cite {newWebCari_Paper} propose a method for caricature identification using CNN and SIFT features extracted from manually labelled facial landmarks. Cross-Modal metric learning method was applied with triplet loss to achieve an accuracy of only 55.48\% on the WebCaricature dataset. 
Cross modal caricature verification can be viewed as a subset of the more generic heterogeneous face matching problem in which the two images to be matched come from different modalities. Techniques for heterogeneous face matching aim to learn a common subspace for the different modalities. He et al., \cite{AAAI_2017} propose to learn modality invariant representations from convolutional networks for near infra-red (NIR) and visual face verification. This model imposes orthogonality constraints for learning the shared and modality specific features. However, it only works for modalities with structural similarity as weights  of the convolutional networks, to extract features from each modality, are tied. Sharing all the weights of convolutional networks that operate on significantly different modalities (such as caricatures) makes it hard for the model to learn shared representations. 


Klare et al., \cite{hfrKlare} propose a method based on kernel similarities to match a probe against a gallery of images of a different modality. They experiment with NIR, visual, thermal and forensic sketches. This method performs well on NIR, visual and thermal images but fails for forensic sketches containing slight distortions to the facial features. Similarly, Crowley et al.,\cite{rw4} used Fisher vector and CNN representations combined with discriminative dimensionality reduction and SVM classifier to search for similar portraits from a gallery given a probe. The portraits consisted of various media like oil, ink, water color and various styles like caricature, pop art, and minimalist. This approach resulted in very low accuracy 26-36\% for the identity retrieval task. 


Huo et al., \cite{ACM_huo_cross_modal} proposed an ensemble of weak cross modal metrics for measuring similarity of facial positions or regions common to both the modalities. This method is not applicable on caricatures as identifying common regions in both the modalities is a hard task and further the same positions would result in unrelated features due to the presence of distortions. 
Shu et al. \cite{ACM_weak_shared} propose a deep network architecture to transfer cross domain information from text to image. The proposed approach relies on strong priors on the weights of the layers shared between the independent modality networks. This prior that the weights of the caricature and visual image networks should be similar, if not tied, is still a very strong assumption, due to the distortions present in caricatures. Our experimental results do suggest that untied weights is a better choice over tied weights between the modalities. Approaches that learn shared subspaces \cite{sharedSubspace} are limited to only learning linear transformations which might not be a good hypothesis especially for our task due to extreme distortions.

\section{Caricature and Visual Images (CaVI) Dataset}

Although there are a few publicly available datasets like the IIIT-CFW dataset \cite{MishraECCV16},  but the number of caricatures in such dataset are very less (about 10) per identity. Due to this reason, our first objective was to create a repository of caricature and visual images. The dataset should reflect the diversity in caricatures of an identity due to different points of view, exaggeration of different facial features, artistic styles, etc.

Visual and caricature images were scrapped through Bing Image Search API \cite{BingSearch}. It was ensured that the search resulted in at least 7 caricature images for every identity. 
Diversity in the caricatures was ensured as only those identities with different types of exaggeration and distortions in the caricatures were included. There were no restrictions on the type of visual images of these identities making the dataset more generic for cross modal face analysis. Our queries to the API for scraping the caricatures were \textit{identity name + Caricature} and for the visual images were \textit{identity name}. The irrelevant images yielded by the API were manually removed. 


The images thus obtained could not be directly used for algorithm development. Often caricatures are not restricted to only the face of a person. Caricatures also encompass the exaggerations of the body features. As our task is focused only on faces, we extracted the faces from the scrapped images. There are many open access face extraction tools available for extracting faces from images \cite{amos2016openface}, \cite{Viola_jones}. We used the popular tool OpenFace\cite{amos2016openface} to extract the bounding box containing faces from visual images. The tool was successful in providing a bounding box for 99.6\% of visual images. We ensured that the obtained results were accurate by manually verifying the estimated bound box.

\begin{figure}[h]
\centering
\includegraphics[width=0.42\textwidth]{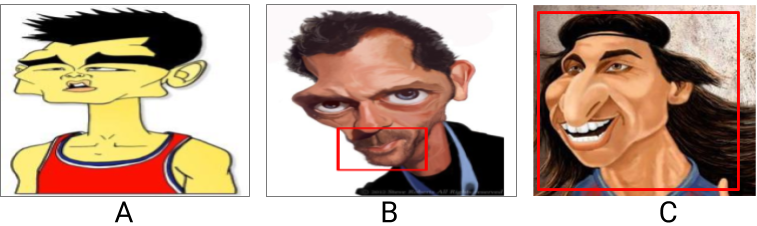}
\caption{Face extraction results with Open Face: A) No face has been detected, B) Face has been detected incorrectly, C) Open Face detected the correct face}
\label{fig:face extraction sample}
\end{figure}

However the tool was unsuccessful in estimating the facial bounding box for caricature images. The bounding box containing the faces were estimated by the tool only for 89.2\% of caricature images. Further not all the estimated regions actually corresponded to the face. For instance, as illustrated in the Figure \ref{fig:face extraction sample}, the estimated bounding box of the face for a caricature image contains only the exaggerated mouth. This is expected as caricatures are distortions of faces. The standard face extraction tool are trained on visual face images, where the faces are of similar geometry. There is no face extraction tool that can reliably extract faces from the caricature images. Thus we manually extracted the faces from the caricatures. During the process, we removed the images that contained profile views of the faces and only annotated those images where the face was completely visible.

Finally the CaVI contains images of 205 identities. There are 5091 caricatures ranging from 10-15 images per identity and 6427 visual images ranging from 10-15 images per identity. A few examples from the dataset are shown in Figure \ref{fig:dataset}.



The CaVI dataset is fundamentally different from the other multi-modal datasets such as \cite{rw3}, \cite{rw1}, \cite{Facenet}. These datasets contain multi-modal images with one-to-one correspondence between images of different modalities that goes beyond the identity in the image. Often the multi-modal images of the person are captured in the same setting, and therefore one can establish a clear one-to-one mapping between the images, which is exploited for algorithm development. In contrast, the CaVI dataset does not have any one-to-one correspondence between the caricature images and real faces (It is impossible to create the one-to-one correspondence) beyond that identity of the person. Thus it is a more challenging dataset.

\section{Methodology}
\begin{figure*}[h]
\centering
\includegraphics[width=\textwidth]{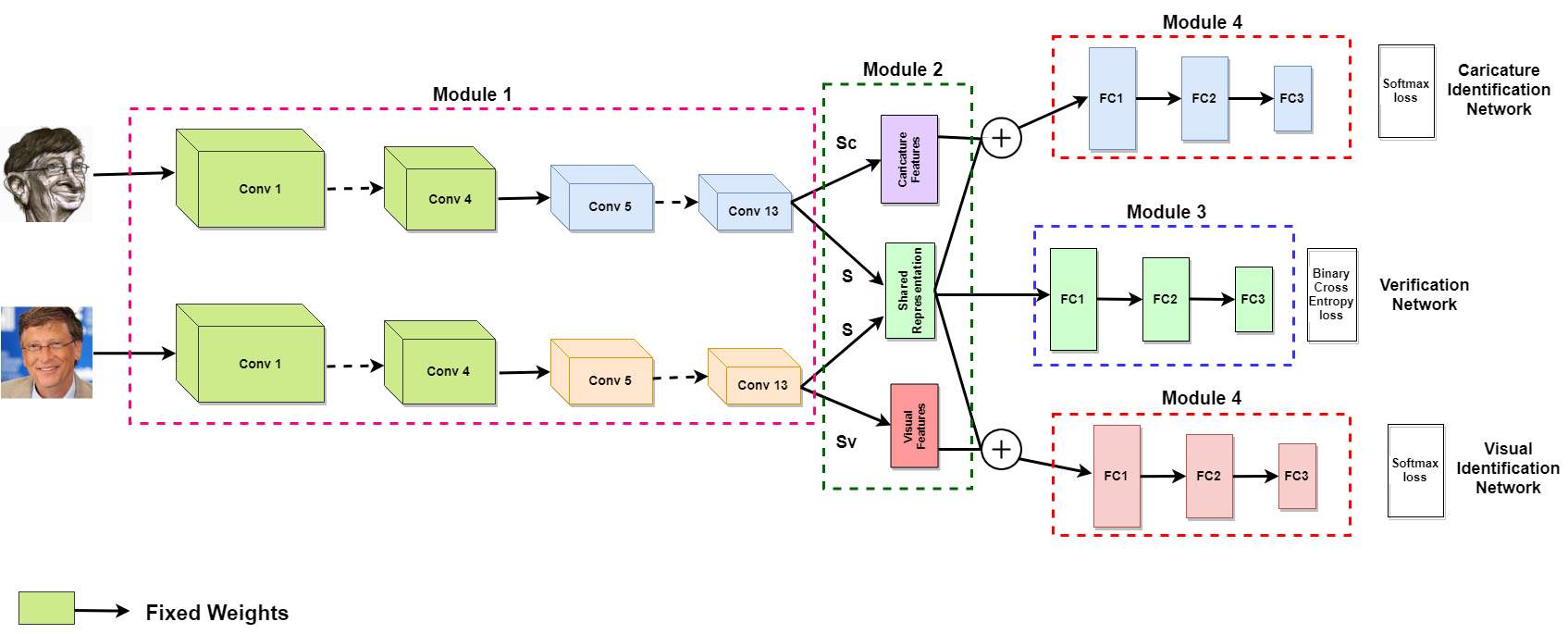}
\caption{
Proposed Architecture for Cross Modal Caricature-Visual Face Verification and Identification}.
\label{fig:architecture} 
\end{figure*}

Let $\{I_{c}\}_{i=1}^{N_c}$ and $\{I_{v}\}_{i=1}^{N_v}$ represent the set of visual and caricature images respectively in the training dataset. The overall architecture of the proposed CaVINet is illustrated in Figure \ref{fig:architecture}. The architecture consists of 4 modules. The first module converts the raw  images from these modalities into rich representations through separate convolutional networks (ConvNet). Let the ConvNet feature representation be denoted by ${ X }_{ i }= f\left( { I }_{ i }; { \Theta }_{ i } \right) $, where $i\in \{ { v },{ c }\} $, $f(.)$ is the ConvNet feature descriptor function that includes multiple layers of convolution, non-linear activation, and pooling, and $\Theta_{i}$ denotes the convolutional parameters for $i^{th}$ modality ConvNet. The parameters of the ConvNet of these two modalities i.e $\Theta_{c}$ and $\Theta_{v}$ are engineered to be different to capture the salient aspects of the images of the two modalities independently, unlike some of the prior approaches \cite{AAAI_2017}. Our objective is to learn transformations that aligns the two feature representations $X_c$ and $X_v$ i.e., captures the common features across the modalities and while filtering out modality specific features.

The second module of the architecture captures the shared and modality specific features in the cross modal representations obtained from the first module. In order to capture the shared and modality specific features, we learn three sets of transformations {$S,S_v,$ and $S_c$} that are applied to the feature representations $X_v$ and $X_c$. The transformation $S$ captures the common features across both the modalities and thus projects the feature representations $X_v$ and $X_c$ onto a common subspace. This common subspace is used to align the representations of images belonging to the same identity from the two modalities. The modality specific features that are important for recognizing the identity in the images are extracted using the transformations $S_v$ and $S_c$. Inspired by Sun et al., \cite{AAAI_2017}, we impose the constraint that the transformations $S$ and $S_c$ (similarly $S$ and $S_v$) must be orthogonal to each other to minimize the redundancy in the features captured by both the transformations.
\begin{equation}
    S_c^T S = 0; \mbox{and }  S_v^T S = 0
\end{equation}
Enforcing the orthogonality constraint ensures that the modality specific features are captured by the transformations $S_c$ and $S_v$, while features shared across the modalities is captured by the transformation $S$. The shared representations across the modalities are defined as $F_c= X_c^T S$ and $F_v = X_v^T S$, and the modality specific representation of the cross modal images is defined as $G_c = X_c^Ts_c$ and $G_v = X_v^T S_v$.

The third module consists of the cross modal verification network that verifies whether the input images from the different modalities correspond to the same identity. This network is learned using the concatenated shared representations $F_c$ and $F_v$. The parameters of the verification network are denoted as $\Phi_{ve}$. Finally, the fourth module consists of the  modality specific identification networks that predict the identity of the person in each of the cross modal images. These networks are trained using the concatenation of the shared and modality specific representations ($[F_c G_c]$ and $[F_v G_v]$) separately for each modality. The parameters of the visual and caricatures identification networks are defined as $\Phi_{ci}$ and $\Phi_{vi}$ respectively. The cross modal verification and modality specific identification networks are trained simultaneously. 
There are two advantages of adding separate identity classification networks. Firstly, it acts as a regularizer to the network  \cite{inception} and secondly, it helps to do both verification and identification in the same model. We demonstrate that including the identification network actually enhances both the verification and the identification accuracy.

The overall loss ($L$) of the CaVINet model, as defined in Equation \ref{eq:loss}, is a weighted sum of three independent loss functions namely, the verification loss ($L_{ve}$, weighted by $\alpha$), the identification loss for caricature images ($L_{ci}$, weighted by $\beta$) and the identification loss for visual images ($L_{vi}$, weighted by $\gamma$).
\begin{equation}
    L= \alpha L_{ve} + \beta L_{ci} + \gamma L_{vi}
    \label{eq:loss}
\end{equation}

The verification loss, $L_{ve}$, is a binary cross entropy loss, where 1 indicates that the identities in the input caricature and visual image pair are the same. If for an input pair of cross-modal images, $y_{ve}$ is the ground truth and $o_{ve}$ is the output predicted by the verification network, then $L_{ve}$ is defined as follows
\begin{equation}
    L_{ve} = y_{ve} \log o_{ve}
\end{equation}
We chose cross-entropy loss over other functions such as triplet loss \cite{Facenet}, or contrasitive loss \cite{contrastive} to avoid the model learning trivial representations. The cross entropy loss is a weaker loss in comparison to the others as it does not strictly require the shared representation for two images with the same identity, but from different modalities, to be identical. Thus, it is easier to optimize compared to the other loss functions. Unlike prior approaches \cite{AAAI_2017}, explicitly modeling the verification task independent of the identification task facilitates verification for images belonging to identities that were not part of the training set. The modality specific identification losses, $L_{ci}$ and $L_{vi}$, are the corresponding softmax losses as defined in the following equations
\begin{equation}
L_{ci}= y_{ci}\log o_{ci}     
\end{equation}
\begin{equation}
L_{vi} = y_{vi}\log o_{vi} 
\end{equation}
where $y_{ci}$ is the true identity of the caricature image and $o_{ci}$ is the identity predicted by the caricature identification network, and $y_{vi}$ is the true identity of the visual image, and $o_{vi}$ is the identity predicted by the visual identification network.

Thus, the overall CaVINet optimization problem can be defined as 
\begin{eqnarray*}
   \min_{\Theta_c,\Theta_v, S, S_c, S_v, \Phi_{ve}, \Phi_{ci}, \Phi_{vi}} L \\
   \mbox{s.t. } S_c^TS=0 \mbox{ and } S_v^TS=0
\end{eqnarray*}

Using Lagrange multipliers $\Lambda_c\geq0$ and $\Lambda_v\geq0$, the constrained minimization problem can be transformed into the following unconstrained problem 
\begin{equation}
   \min_{\Theta_c,\Theta_v, S, S_c, S_v, \Phi_{ve}, \Phi_{ci}, \Phi_{vi}} L +\Lambda_c\|S_c^TS\|^2+\Lambda_v\|S_v^TS\|^2
   \label{eq:objective}
\end{equation}

We have chosen $\Lambda_{c}=\Lambda_{v}=\Lambda$ to reduce the number of parameters to be tuned and have performed ablations on different values of $\Lambda$ to obtain the optimal value.

The parameters of CaVINet are updated using gradient descent technique. The weight update for the paramters $\Theta_c,\Theta_v, \Phi_{ve}, \Phi_{ci},$ and $\Phi_{vi}$ are straight forward. The derivative of the objective function in Equation \ref{eq:objective} wrt to parameters $S$, $S_{v}$, and $S_{c}$ has additional terms due to the constraints and can be obtained as follows:
\begin{eqnarray*}
\frac{\partial L}{\partial S} & = & \alpha \frac{\partial L_{ve}}{\partial S}+\beta \frac{\partial L_{ci}}{\partial S}+\gamma \frac{\partial L_{vi}}{\partial S}+ 2\Lambda_c S_c S_c^T S+2\Lambda_v S_v S_v^T S \\
\frac{\partial L}{\partial S_c} & = & \beta \frac{\partial L_{ci}}{\partial S} + 2\Lambda_c SS^T S_c \\
\frac{\partial L}{\partial S_v} & = & \gamma \frac{\partial L_{vi}}{\partial S} + 2\Lambda_c SS^T S_v
\end{eqnarray*}

The choice of the number of layers for the CaVINet architecture is presented in Figure \ref{fig:architecture}. Module 1, obtains rich independent representations for the two modalities, consists of 13 convolutional layers of the VGGFace architecture \cite{rw4}. The input to both the streams is images of size $224\times224\times3$. All the convolutional layers are initialized with VGGFace pre-trained weights on LFW \cite{LFWTech}, and Youtube Faces Dataset \cite{youtube} both of which comprise of the visual images. The initial 4 convolutional layers for both the modalities in module 1 are frozen due to the common observation that the low level features across vision tasks remain the same \cite{Zeiler14visualizingand}. The rest of the convolutional layers in module 1 are fine tuned. This is followed by the transformations $S, S_c,$ and $S_v$. Each of the verification, caricature and visual identification networks have a set of 3 fully-connected layers with binary cross-entropy and softmax losses.

We used mini batch stochastic gradient descent as our optimizer with a learning rate of $\eta = 10^{-3}$,  $\emph{decay}=10^{-6}$ and $\emph{batch size}= 25$. In order to make the network robust to small transformations such as translation, rotation, noise etc., and to avoid over fitting, we performed data augmentation. This also helped to increase the training set to $401,769$ cross modal pairs. A cross modal pair consists of one image each from the caricature and visual modalities. If both the images in a cross modal pair contain the face of the same identity, the pair is treated as a positive sample for the verification task (else as a negative sample). The individual identities of the faces in the cross modal pair are treated as the class labels for the corresponding identification tasks. We used a Dropout \cite{dropout_sri} ratio of 0.6 for all the fully-connected layers. We set the value of the Lagrange multipliers to  $\Lambda_{C}=\Lambda_{V}=0.2$ through experimentation.

\section{Experiments and Results}
In this section, we present and discuss the results for various experiments that we have performed to investigate the CaVINet model. This section is divided into three parts: baseline experiments that discusses the results of CaVINet in the context of other state-of-the-art approaches; ablation studies  conducted on CaVINet; and finally, qualitative visual analysis of CaVINet.

We have performed all the experiments on our \emph{CaVI Dataset}. The entire dataset containing images from 205 identities has been divided  into 2 sets: The first set consists of the images from 195 identities and the images from the remaining 10 identities constitute the \emph{Unseen Test Set}. This test set is used to test the performance of the verification model on unseen identities. The first set of images is further divided into mutually exclusive train, test and validation sets. It was ensured that all the identities are present in the train, test and validation sets but the images were not repeated in any of the sets. We call this test set as the \emph{Seen Test Set} as the identities in the test set are also present in the training set.

\subsection{Baseline Experiments}

We compare the performance of CaVINet against other baseline approaches for both the verification and identification tasks.

\subsubsection{Verification Baselines}
\begin{figure}[h]
\centering
\includegraphics[width=0.42\textwidth]{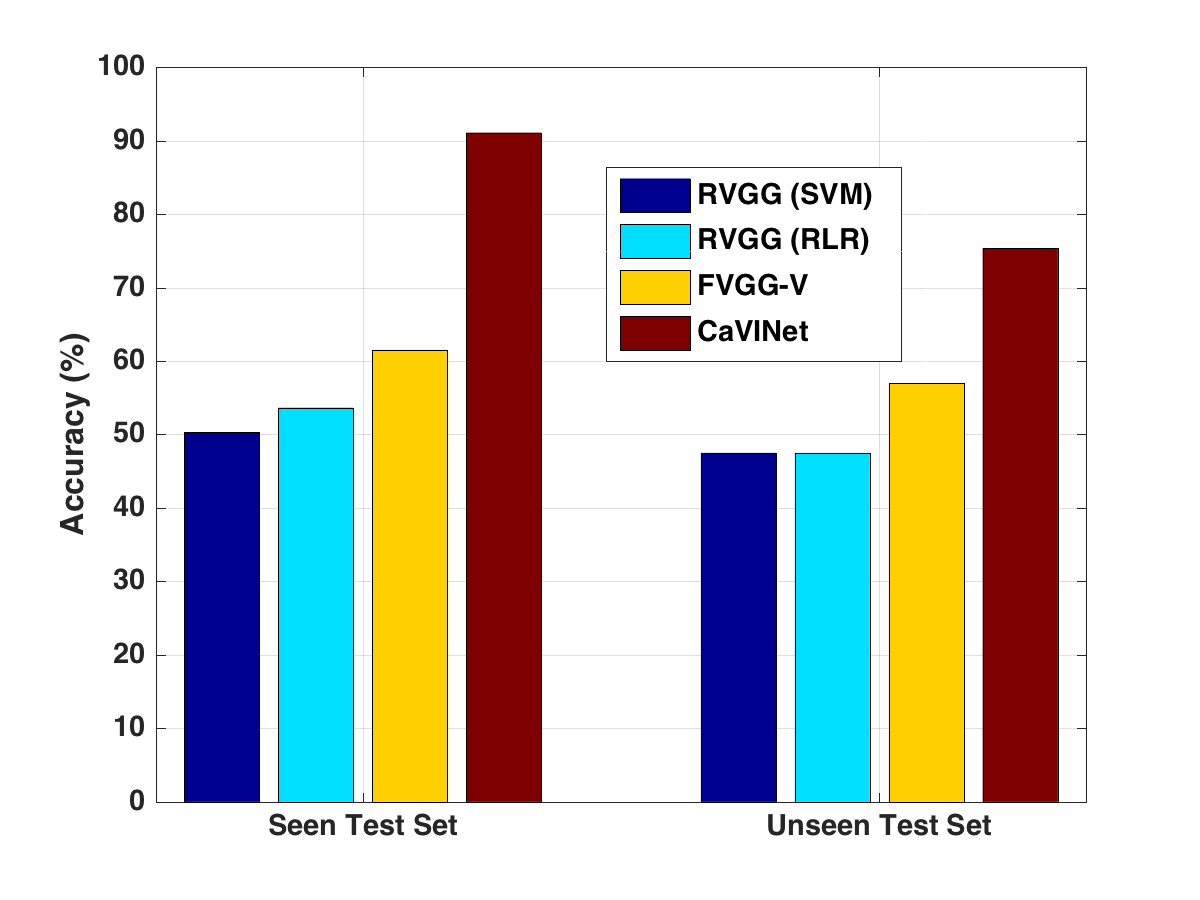}
\caption{[Best Viewed in Color] Verification accuracy on the seen and unseen test sets.}
\label{fig:verification_baseline}
\end{figure}

Two commonly used baselines as described below were used to compare against CaVINet.
\begin{itemize}
  \item RVGG - The off-the-shelf VGG-Face model was used to extract features from the caricature and visual images. The images were represented using the feature maps obtained from the last convolutional layer of the VGG-Face model. These features, for a pair of visual and caricature images, were concatenated to train a binary classifier for cross-modal verification. Support Vector Machine (SVM) \cite{svm} and Regularized Logistic Regression(RLR) were used as the classification models. The optimal hyper-parameters for the SVM and the RLR were obtained through a cross validation process on the training set. The optimal kernel was radial basis kernel and the box penalty parameter was 1000. The optimal regularization parameter value for RLR was $10^{-4}$.

\item FVGG-V (Fine Tuned VGG-Verification) - This is a coupled network with 13 convolutional layers, with tied weights, for processing visual and caricature images that outputs 1 if the input image pair contains the same identity. The weights of the network were initialized using the VGG-Face model, and were fine tuned (except for the first 4 low level layers) using the CaVI training dataset.

\end{itemize}
It is evident from the Figure \ref{fig:verification_baseline} that CaVINet significantly out performs all the other approaches. The performance of RVGG is significantly inferior to that of FVGG-V indicating that the features learned for visual images (as obtained from the pre-trained VGG-Face model) are insufficient to characterize the caricature images. The performance improves when we fine-tune the model using the caricature images as seen in the FVGG-V model. However, the fine tuned model still does not match up to the performance of CaVINet that better characterizes the shared representation between the modalities. The accuracy of the CaVINet model on the \emph{Unseen Test Set} containing images from 10 unseen identities was 75\%. Even though the performance drops considerably in comparison against the set whose identities were part of the training set, it still outperforms the other models significantly. 

\subsubsection{Identification Baseline}
\begin{figure}[h]
\centering
\includegraphics[width=0.42\textwidth]{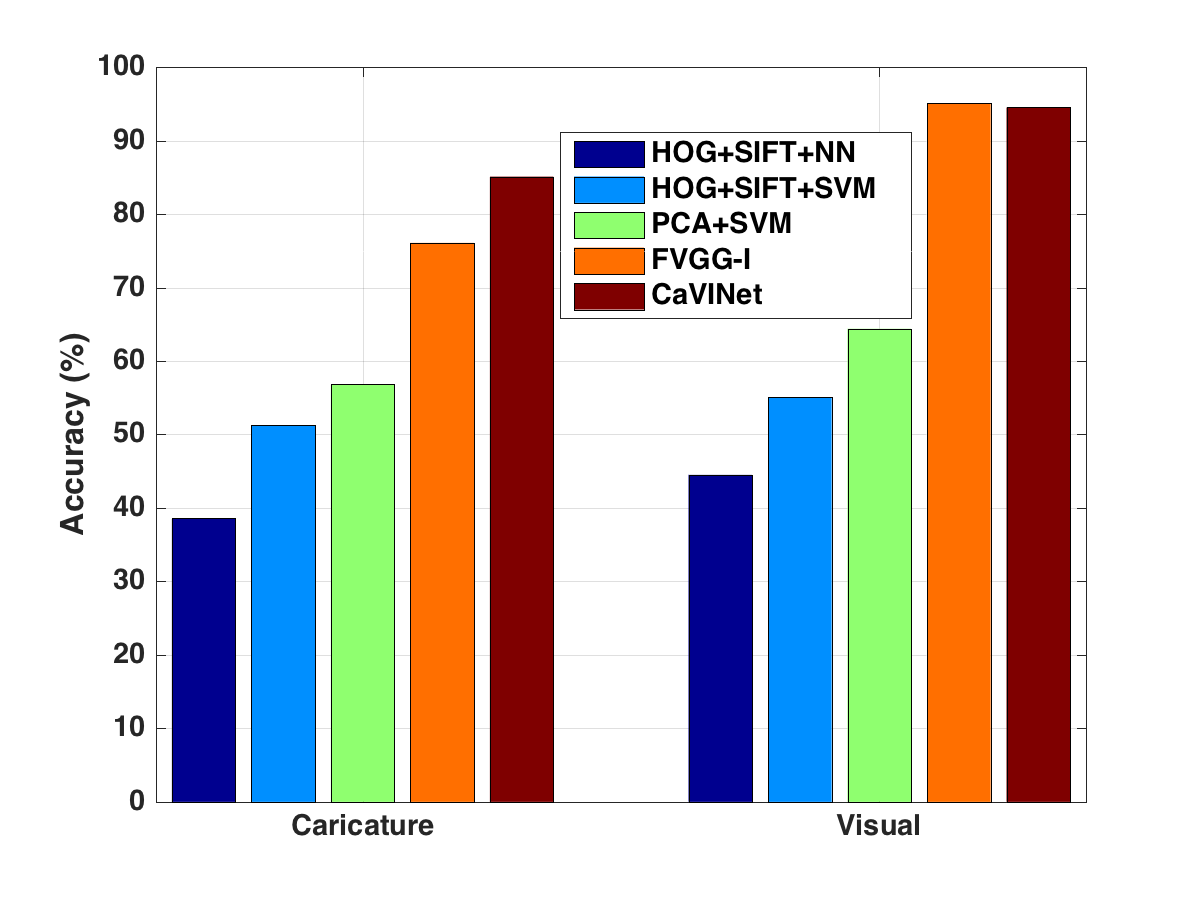}
\caption{[Best Viewed in Color] Identification accuracy on the caricature and visual images.}
\label{fig:identification}
\end{figure}

We compared CaVINet against the following three approaches for the identification task. Rank-1 accuracy was used to measure the performance of the different models.
\begin{itemize}
  \item FVGG-I (Fine tuned VGG-Identification) - Huo et al. \cite{newWebCari_Paper}, utilize off-the shelf VGG-Face model to extract the features at various landmark points for caricature images. These features are used for training models for identification task. We also took a similar approach where, two separate networks were trained for recognizing the identities in the images from the two modalities. These are completely independent networks that do not take advantage of potential knowledge transfer between the modalities. Each network was initialized using the VGG-Face model. Only the last three fully connected layers of the network were fine tuned separately for caricature and visual images. We fine tuned only the fully connected layers, as the size of our image dataset is relatively small compared to the size of the dataset that was used to train the VGG-Face model \cite{Parkhi15}.

  \item PCA - We used the popular principal component analysis based approach for face recognition to test its performance on the caricature dataset. The caricature images were projected onto a lower dimensional subspace (1500 dimensions) using PCA. The projected data was used to train a support vector machine (PCA+SVM) classifier. The optimal kernel choice was linear kernel and box penalty parameter for SVM was 5 which was obtained through experimentation.

  \item HOG \& SIFT - We extracted histogram of oriented gradients (HOG) \cite{hog} and scale invariant feature transform (SIFT) features for the caricature images. Face recognition approaches extract SIFT features only from the facial key points \cite{rw5, face_sift1, face_sift2}. As the CaVI dataset does not have facial key points, we used visual bag of words \cite{visual_bow} of SIFT features. HOG (dimension 4704) and SIFT (dimension 1000) features were concatenated upon which two classifiers, multi-layer perceptron(MLP) and support vector machine (SVM), were trained. The baseline MLP had three layers 5704 (input) $\rightarrow$ 2048 $\rightarrow$ 512 $\rightarrow$ 195 with sigmoid non-linearity.

\end{itemize}
The results for the identification tasks are presented in Figure \ref{fig:identification}. Predictably, the best accuracy among the baselines is obtained by the deep network model - FVGG-I. CaVINet is able to significantly better the performance of FVGG-I for caricature recognition. CaVINet allows for  knowledge transfer from the visual to caricature modality. As a result it is able to learn the identity specific facial features, while ignoring the exaggerations and distortions. Both CaVINet and FVGG-I yield an accuracy of around 95\% on the visual images, which is less than some of the state-of-the-art models  for face recognition by about 5\%. We attribute this to the significantly smaller number of training images per identity present in the CaVI dataset.
\subsection{Ablations on CaVINet}

We performed various ablations on CaVINet to investigate the design choices that were made for engineering the architecture.
The verification and identification accuracy, as discussed earlier, suggests that cross modal verification task is a harder task than the modality specific identification tasks. This motivated us to vary the weights $\alpha$, $\beta$ and $\gamma$ of independent loss functions. The results for this ablation study are summarized in Table \ref{tab:lossfunctionweights}. We observed that the ratio 55:30:15 for $\alpha:\beta:\gamma$ results in the best performance. The performance of the model with uniform weights to all the loss functions is significantly poorer indicating the merit of increasing the emphasis on the loss for the verification task. 
\begin{table}[!h]
  \begin{center}
    
    \scalebox{0.92}{
    \begin{tabular}{|c|c|c|c|} 
    \hline
       $\alpha:\beta:\gamma$ & \textbf{Verification} & \textbf{Visual-Id} & \textbf{Caricature-Id}\\
      \hline
      55:30:15 & \textbf{91.06} & \textbf{94.50} & \textbf{85.09} \\ \hline 
      50:25:25 & 86.31 & 93.34 & 81.02  \\ \hline 
      40:35:25 & 83.46 & 80.96 & 84.02 \\ \hline 
      33:33:33 & 79.43 & 92.74 & 82.64 \\ \hline 
    \end{tabular}}
    \bigskip
    \caption{Ablation Study on the optimal ratio of the loss function weights $\alpha:\beta:\gamma$.}
    \label{tab:lossfunctionweights}
  \end{center}
\end{table}

We explored the choice of untied weights in module 1 of CaVINet. We experimented with shared weights between all the layers of the caricature and visual modality in module 1 similar to the approach suggested by  Shu et al., \cite{ACM_weak_shared} Using tied weights reduces the number of parameters to be learned by the model, and thereby helps to improve the generalization performance when training data is scarce. However, the results presented in Table \ref{table1} show a significant drop in the performance for the verification and visual identification tasks. Interestingly there is a marginal increase in the performance on the caricature identification task. This perhaps, can be attributed due to the manner in which the tied weight model represents the images of two significantly different modalities. The caricature and visual modalities have little structural similarities (in contrast to near infra-red and visual images) that makes it difficult for a model to obtain identical characterization of both the modalities using tied weights as indicated by the significant drop in both verification and visual identification tasks.

We also investigated the efficacy of the orthogonality constraints for learning the shared and modality specific features. Removing the orthogonality constraints would result in the model learning only shared representations between the modalities. The resulting shared representation was used for training the verification and identification models. As can be seen from Table \ref{table1}, there is a significant drop in the performance of the model on all tasks. The maximum performance drop is observed on caricature identification task. This suggests the utility of modality specific features, captured by the transformation $S_c$, for learning the identities of the faces. We also conducted an experiment where, while retaining the orthogonality constraint, only the shared representations were used for both the verification and identification tasks. Here too we observed a drop of 5\% in the performance on both identification tasks further strengthening the need for modality specific features. We also used only the Visual features to do the identification task and saw a drop in the accuracy which indicates that both modality specific and shared features are important for the task.

\begin{table}[!h]
  \begin{center}
    
    \scalebox{0.92}{
    \begin{tabular}{|l|c|c|c|} 
    \hline
      \textbf{Experiment} & \textbf{Verification} & \textbf{Visual-Id}& \textbf{Cari-Id}\\
      \hline
      CaVINet  & \textbf{91.06} & \textbf{94.50} & 85.09 \\ \hline 
      CaVINet(TW) & 84.32 & 85.16 & \textbf{86.02}\\ \hline
      CaVINet(w/o ortho)  & 86.01 & 93.46 & 80.43\\ \hline
      CaVINet(shared features)  & 88.59 & 90.56 & 81.23\\ \hline
      CaVINet(Visual features)  & 88.58 & 92.16 & 83.36\\ \hline
    \end{tabular}}
    \bigskip
    \caption{Ablation study on CaVINet with tied weights (TW), without the orthogonality constraints (w/o ortho) and learning only from the shared representations (shared features).}
    \label{table1}
  \end{center}
\end{table}

\begin{figure}[h]
\centering
\includegraphics[width=0.42\textwidth]{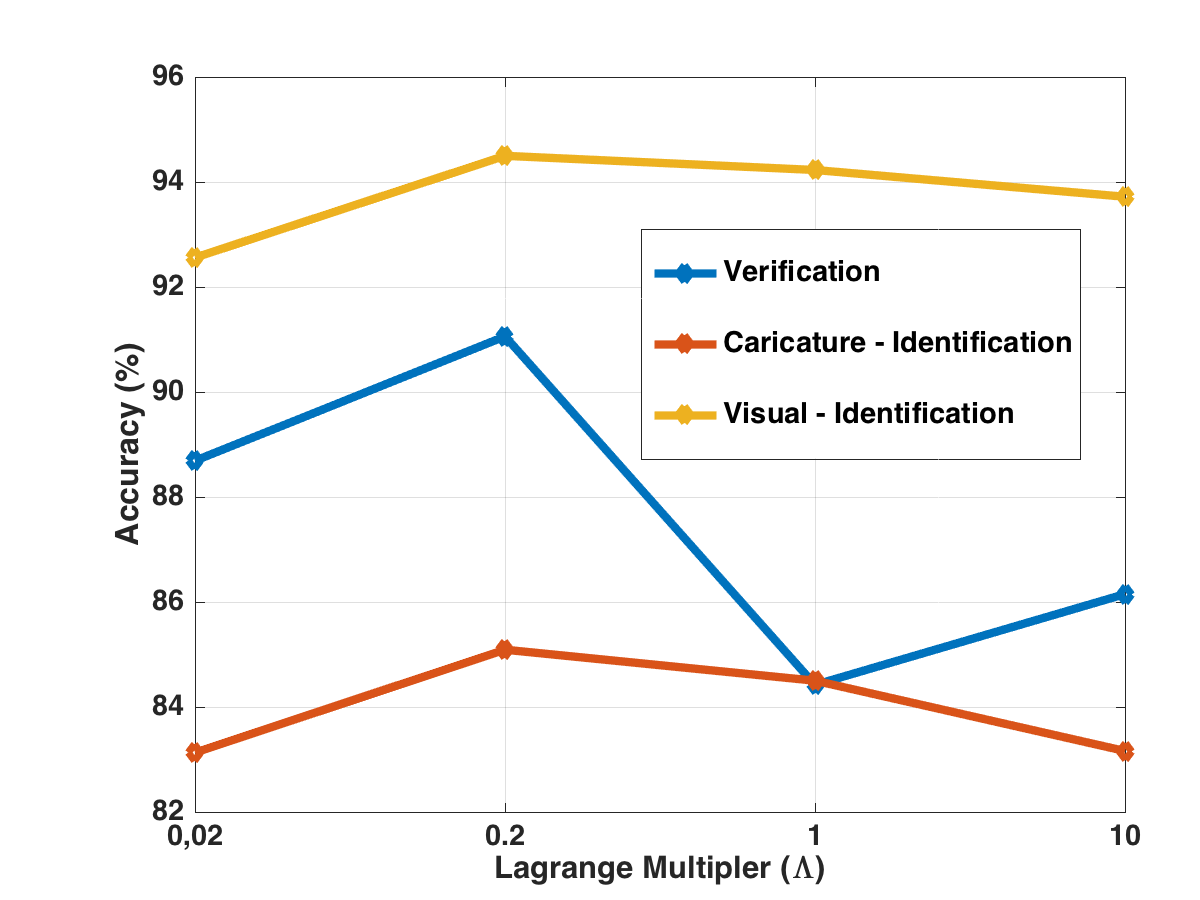}
\caption{[Best Viewed in Color] Ablation study on values for Lagrange multiplier $\Lambda$. The optimal value was obtained at $\Lambda=0.2$.}
\label{fig:lambda}
\end{figure}

The Lagrange multipliers associated with the orthogonality constraints are parameters used to indicate the extent to which the constraints have to be satisfied. Setting the value to 0 would result in having no orthogonal transformations, and larger values indicate more penalty to the model for learning non-orthogonal transformations. We varied the value of the multipliers to study its effect on model performance. The results of this study are presented in Figure \ref{fig:lambda}.  As evident from the results, the best performance was obtained for $\Lambda = 0.2$, indicating that enforcing the orthogonality constraint helps in improving the performance of CaVINet.

\subsection{Qualitative Visual Analysis}
\begin{figure}[h]
\centering
\includegraphics[scale=0.35]{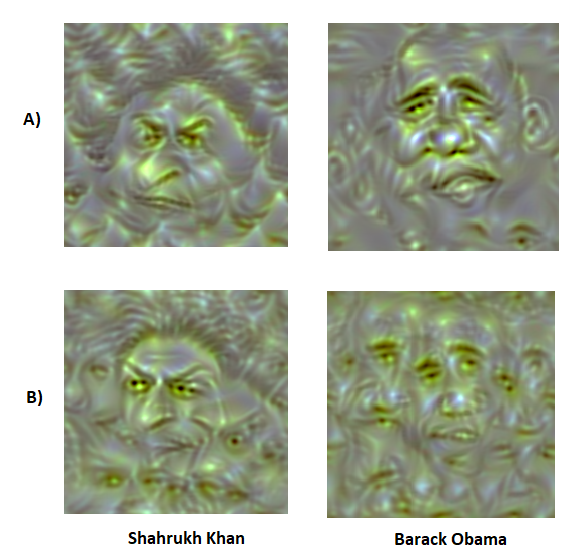}
\caption{A) Activation maximization on softmax layer neurons for Shahrukh Khan and Barack Obama of A) caricature and B) visual identification networks respectively.}
\label{fig:Activation maximization}
\end{figure}

We employed activation maximization and saliency map based visualization techniques to qualitatively understand the CaVINet model. We synthesized preferred input images through activation maximization constrained on the natural image prior of jitter. \cite{vis4, vis6}. Figure \ref{fig:Activation maximization} presents the synthesized input images when the activation of the neurons corresponding to the softmax layer of the caricature (and visual) identification network is maximized. It can be observed that caricature network does learn some of the variations and distortions present in caricatures, along with the modality invariant features of a person's face. The modality invariant facial features are also present in the images synthesized for the visual identification network; however the distortions are absent. Further, both the synthesized images are rotation and flipping invariant.

\begin{figure}[h]
\centering
\includegraphics[scale=0.3]{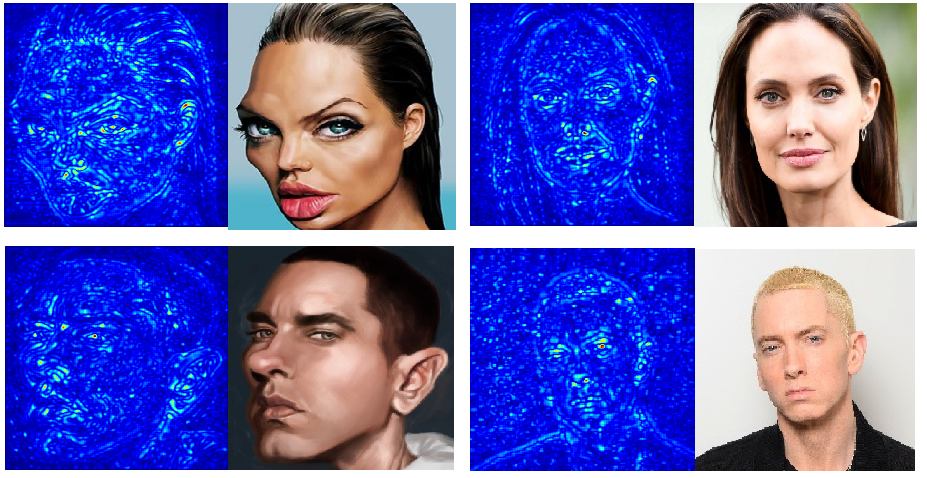}
\caption{[Best viewed in Color] Visualizing rectified gradient output of shared + unique layer of caricature and visual network respectively for sample inputs Angelina Jolie and Eminem.}
\label{fig:Saliency map}
\end{figure}

The saliency maps based visualization technique \cite{Zeiler14visualizingand}, \cite{vis2} was used to study the features in the input caricature and visual images that were important for the identification model. In particular, rectified saliency was applied on the shared and modality specific features of our model for visualizing the positive gradients. As this is an unsupervised visualization technique, the visualized images show the features that are used by both the shared and modality specific layers combined. It can be seen from figure \ref{fig:Saliency map}, on a caricature and a visual image of Angelina Jolie and Eminem, the gradient output is much higher for specific facial features that have striking similarity in both the caricature and visual images.

\begin{figure}[!h]
\centering
\includegraphics[scale=0.15]{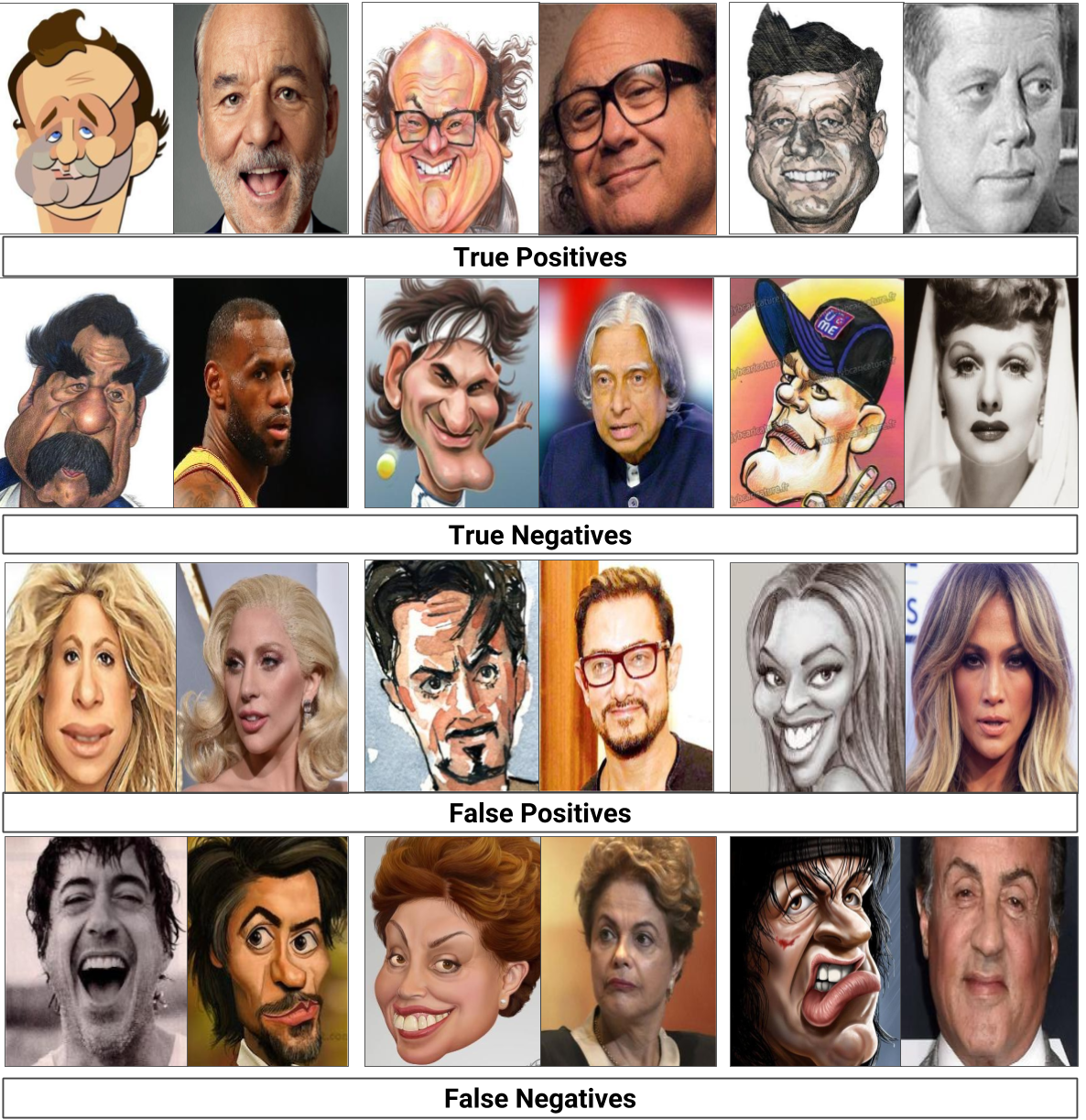}
\caption{Some examples of True Positives, True Negatives, False Positives and False Negatives of CaVINet verification model.}
\label{fig:examples}
\vspace{-2mm}
\end{figure}

Figure \ref{fig:examples} presents examples from the confusion matrix of the CaVINet verification. It is evident from the true positives and negatives that the model is able to predict accurately in the presence of distortions in the caricatures. It is interesting to note that the set of false positives predicted by the verification model have significant similarities among the images of the two modalities. It is a difficult task even for a human to identify some of these image pairs as belonging to different identities. The set of false negatives suggests that the model is unable predict accurately when the distortions in the caricature are quite different from the view of the visual image. For instance in the left most image pair, face in the visual image has an open mouth, while the caricature has a closed mouth. In the right most pair, the caricature has significant distortions. In both the examples either the caricature or the visual image has the face partially occluded. The results suggest that there is still scope for improving the current model to handle these distortions.

\section{Summary}
This paper presents the first cross modal architecture that is able to handle extreme distortions present in caricatures for verification and identification tasks. It introduces a new publicly available large dataset containing 5091 caricatures and 6427 visual images from 205 identities, along with bounding box of the faces. It proposes a novel coupled deep network - CaVINet for performing the verification and identification tasks simultaneously. 
The coupled architecture of the model that bridges caricature and visual modality facilitates successful transfer of information across the modalities. This is demonstrated in our experiments where the CaVINet model achieves 91\% accuracy for verifying caricatures against visual images and 85\% accuracy on the challenging task of recognizing the identities in the caricature images. CaVINet overcomes the bottleneck of prior cross-modal verification approaches that require the identity of a test image to be present during training by explicitly learning a verification network. 

\begin{acks}
 The authors gratefully acknowledge NVIDIA for the hardware grant. This research is supported by the Department of Science and Technology, India under grant YSS/2015/001206.
\end{acks}

\bibliographystyle{ACM-Reference-Format}
\balance
\bibliography{cavinet} 

\end{document}